\documentclass[11pt]{article}

\usepackage[preprint]{acl}

\usepackage{times}
\usepackage{latexsym}

\usepackage[T1]{fontenc}

\usepackage[utf8]{inputenc}

\usepackage{microtype}

\usepackage{inconsolata}

\usepackage{graphicx}

\usepackage{amsmath} 
\usepackage{amsfonts}
\usepackage{multirow} 
\usepackage{graphicx}
\usepackage{subcaption}

\title{LLM-Guided Knowledge Distillation for Temporal Knowledge Graph Reasoning}

\author{Wang Xing\textsuperscript{1},
Wei Song\textsuperscript{2}, 
Siyu Lin\textsuperscript{3},
Chen Wu\textsuperscript{2},  
Man Wang\textsuperscript{2} \\
$^1$School of Computer Science and Technology, Xidian University\\ 
$^2$School of Computing and Artificial Intelligence, Southwest Jiaotong University\\
$^3$School of Information Science and Engineering, Chongqing Jiaotong University\\
}

\begin{document}
\maketitle
\begin{abstract}
Temporal knowledge graphs (TKGs) support reasoning over time-evolving facts, yet state-of-the-art models are often computationally heavy and costly to deploy. Existing compression and distillation techniques are largely designed for static graphs; directly applying them to temporal settings may overlook time-dependent interactions and lead to performance degradation.
We propose an LLM-assisted distillation framework specifically designed for temporal knowledge graph reasoning. Beyond a conventional high-capacity temporal teacher, we incorporate a large language model as an auxiliary instructor to provide enriched supervision. The LLM supplies broad background knowledge and temporally informed signals, enabling a lightweight student to better model event dynamics without increasing inference-time complexity. Training is conducted by jointly optimizing supervised and distillation objectives, using a staged alignment strategy to progressively integrate guidance from both teachers.
Extensive experiments on multiple public TKG benchmarks with diverse backbone architectures demonstrate that the proposed approach consistently improves link prediction performance over strong distillation baselines, while maintaining a compact and efficient student model. The results highlight the potential of large language models as effective teachers for transferring temporal reasoning capability to resource-efficient TKG systems.

\end{abstract}

\section{Introduction}

Temporal knowledge graphs (TKGs) attach timestamps to facts and enable modeling how relations change over time \citep{leblay2018deriving, kazemi2020representation}. They underpin many applications, including monitoring system states, semantic retrieval, and planning \citep{garcia2020temporal}. Popular benchmarks include YAGO \citep{mahdisoltani2015yago3} and WIKI/Wikidata \citep{vrandevcic2014wikidata}. Before temporal settings, embedding-based models for static KGs established strong link-prediction baselines \citep{bordes2013translating}. Methods like TransE represent relations as translations \citep{bordes2013translating}, RotatE uses complex-space rotations \citep{sun2019rotate}, ComplEx relies on complex bilinear scoring \citep{trouillon2016complex}, and SimplE improves expressiveness with paired embeddings \citep{kazemi2018simple}. These ideas are frequently reused when building time-aware variants.

For TKGs, early methods extended translation-style scoring with temporal components \citep{leblay2018deriving}, while others employed neural architectures (e.g., recurrent or convolutional modules) to aggregate histories \citep{jin2020recurrent}. Nonetheless, naively adapting static techniques can under-model event dynamics and time-dependent constraints. Recent work has explored incorporating explicit temporal logical structures to enhance reasoning over temporal knowledge graphs. Differentiable logical rule learning frameworks \citep{xiongtilp, xiong2024teilp} model temporal dependencies through logical rules, enabling interpretable temporal reasoning and time prediction.

Most evaluations consider two regimes: interpolation (completion within an observed window) and extrapolation (forecasting beyond the latest time) \citep{xu2020temporal}. HyTE is a representative interpolation approach \citep{dasgupta2018hyte}, while extrapolation has been studied via continuous-time or history-driven models \citep{trivedi2017know}. Examples include Chronos \citep{sadeghian2021chronos}, Know-Evolve \citep{trivedi2017know}, ExTTE \citep{xu2020temporal}, CyGNet \citep{zhu2021cygnet}, and RE-NET \citep{jin2020recurrent}. While effective, many such models become large, increasing training and deployment cost. This motivates efficient TKG reasoning, as real systems often need low-latency inference under strict memory and power budgets \citep{garcia2020temporal}. Model compression and knowledge distillation are natural options \citep{hinton2015distilling}.

Model compression and inference acceleration have attracted increasing attention in recent years. Knowledge distillation has proven effective across domains such as computer vision and recommendation systems \citep{hinton2015distilling}. Within knowledge graph research, distillation techniques are commonly used to transfer representations from large teachers to lightweight students. However, most existing approaches are designed for static knowledge graphs and rarely account for temporal dynamics \citep{zhang2020relational}.

Large language models (LLMs), including GPT-3 \citep{brown2020language}, PaLM \citep{chowdhery2022palm}, and LLaMA \citep{touvron2023llama}, exhibit strong reasoning and generalization abilities derived from large-scale pretraining. Techniques such as instruction tuning and reinforcement learning from human feedback further enhance their decision-making capabilities \citep{ouyang2022training}. Recent findings indicate that LLMs can handle complex reasoning tasks and generalize to diverse real-world scenarios \citep{xiong2024large, yang2024can}. Moreover, they are capable of structured reasoning and deliberate planning through implicit world-model mechanisms \citep{yang2024harnessing, xiong2025deliberate}. These characteristics make LLMs suitable as powerful teacher models in a distillation framework.

Inspired by these insights, we introduce a distillation strategy specifically designed for temporal knowledge graph reasoning. The proposed framework employs LLMs as auxiliary teachers to inject both background knowledge and time-aware reasoning signals into the training process. By supervising compact student models with enriched temporal semantics, our method facilitates efficient deployment in resource-constrained environments. Experimental evidence shows that the proposed approach surpasses conventional temporal KG models while substantially reducing computational and storage overhead.

\begin{figure*}[t]
\centering
\begin{subfigure}[t]{0.4\textwidth}
  \centering
  \includegraphics[width=\linewidth]{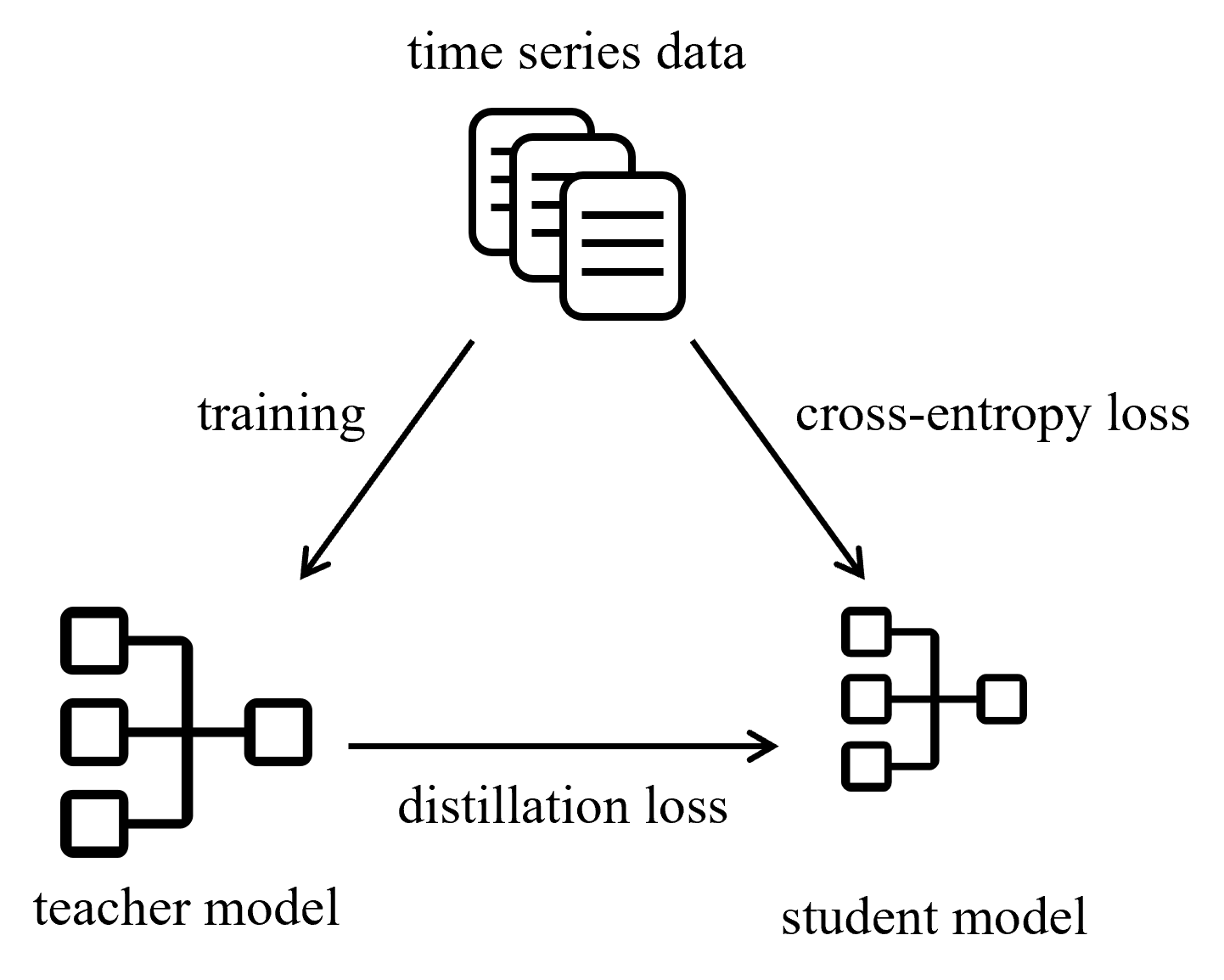}
  \caption{Traditional knowledge distillation method}
\end{subfigure}
\hspace{2pt}
\begin{subfigure}[t]{0.4\textwidth}
  \centering
  \includegraphics[width=\linewidth]{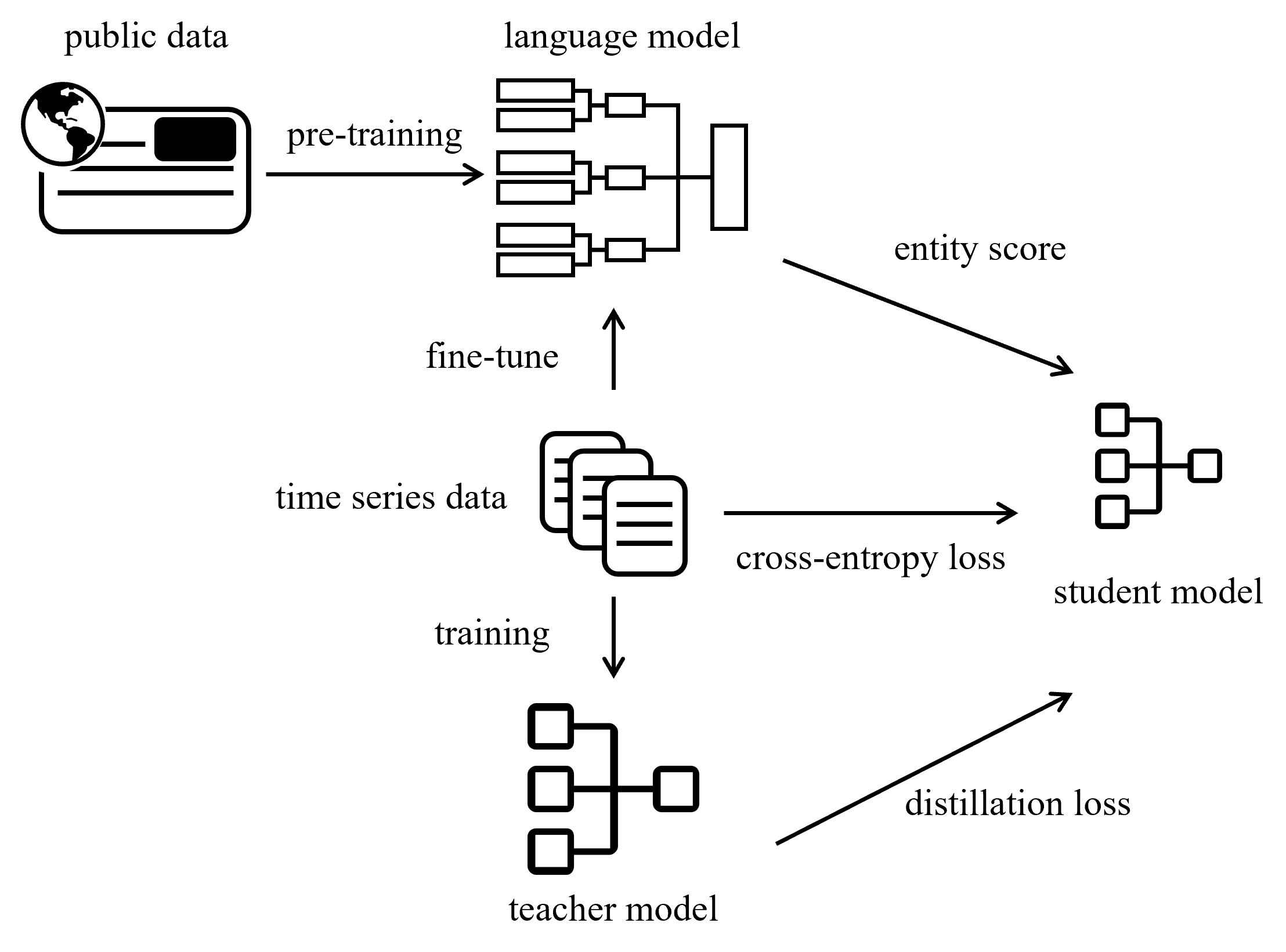}
  \caption{Large language model based distillation method}
\end{subfigure}
\caption{Comparison between different distillation methods.} 
\label{fig:distillation_comparison}
\end{figure*}

\section{Distillation for Temporal Knowledge Graph Reasoning Models Enhanced by LLMs}

\subsection{Problem Definition}

We represent a temporal knowledge graph (TKG) as timestamped quadruples \( g = (s, p, o, t) \), where \( s, o \in E \) are entities, \( p \in R \) is a relation, and \( t \in T \) is the timestamp. The TKG can be structured as a sequence of snapshots \( \mathcal{G} = \{ \mathcal{G}_1, \ldots, \mathcal{G}_T \} \), with \( \mathcal{G}_t \) containing all observed facts at time \( t \).

The task focuses on time-conditioned link prediction: given queries like \( (s, p, ?, t) \) or \( (?, p, o, t) \), the model ranks candidate entities to provide the most plausible completion. In extrapolation tasks, queries may extend beyond the final training snapshot, requiring generalization from historical patterns to future time points.

At time \( t \), the task involves constructing incomplete fact quadruples \( \mathcal{G}_t' \) by replacing the subject or object entity in the original set \( \mathcal{G}_t \). These incomplete fact sets, combined with the original facts, form the complete temporal graph \( \mathcal{G}_c \), defined as
\begin{equation} 
\mathcal{G}_c = \mathcal{G} \cup \mathcal{G}'.
\end{equation}
The incomplete facts are:
\begin{equation} 
\begin{aligned} 
\mathcal{G}' =\;& \big\{(s', p, o, t) \mid s' \in E, s' \neq s\big\} \\ &\cup \big\{(s, p, o', t) \mid o' \in E, o' \neq o\big\}. 
\end{aligned} 
\end{equation}

\subsection{Model Overview}

We distill temporal reasoning into a compact student model \( S \), guided by two teachers: a task teacher \( T \), a high-capacity TKG model trained on \( \mathcal{G} \), and a large language model (LLM) \( M \). The task teacher provides robust supervision, while the LLM contributes complementary semantic knowledge, especially for sparse or ambiguous temporal data.

Training involves two phases. In Phase 1, the student model \( S \) aligns with \( T \) to learn a stable temporal scoring function. Phase 2 incorporates guidance from \( M \) to further refine the student's ranking behavior and improve generalization. At inference, only the student model is used, ensuring low deployment cost.

\subsection{Loss Functions}

Three loss functions are used during training, combined into the final objective function. Let \( L_1 \) be the distillation loss between the student and the teacher model, \( L_2 \) be the distillation loss between the student and the LLM, and \( L_3 \) be the supervised loss based on ground truth labels. The total loss is:
\begin{equation}
L_{\text{total}} = L_1 + L_2 + \beta L_3.
\end{equation}

The first loss function, \( L_1 \), transfers temporal reasoning knowledge from the teacher model to the student using an encoder-decoder architecture. The distillation loss is defined as:
\begin{align}
L_1 =\;& \alpha \, \mathcal{C}\big(g \log f_T(s, p, o, t)\big) \nonumber \\
& + (1 - \alpha)\, \mathcal{C}\big(g \log f_S(s, p, o, t)\big).
\end{align}
where \( f_T \) and \( f_S \) are the teacher and student scoring functions, respectively, and \( \mathcal{C}(\cdot) \) denotes the softmax function.

The second loss, \( L_2 \), minimizes the discrepancy between the student model and LLM predictions using the Huber loss:
\begin{equation}
L_2 = 
\begin{cases}
\frac{1}{2} (f_T - f_S)^2, & |f_T - f_S| \le \delta, \\
\delta |f_T - f_S| - \frac{1}{2} \delta^2, & |f_T - f_S| > \delta,
\end{cases}
\end{equation}
where \( f_T \) and \( f_S \) are the prediction scores from the LLM and student, and \( \delta \) is a threshold hyperparameter.

The third loss, \( L_3 \), is the supervised loss based on ground truth labels. It uses the LLM to encode entity relations, producing vector representations. The loss is computed as the mean squared error:
\begin{equation}
L_3 = \| f_S(p) - P(p) \|_2^2,
\end{equation}
where \( f_S(p) \) is the student model's prediction and \( P(p) \) is the ground truth label.

The final objective combines these losses:
\begin{equation}
L_{\text{total}} = L_1 + \alpha L_2 + \beta L^{\text{LLM}}_3,
\end{equation}
where \( L^{\text{LLM}}_3 \) is the LLM-assisted distillation loss.

\begin{figure*}[t]
  \centering
  \includegraphics[width=0.9\linewidth]{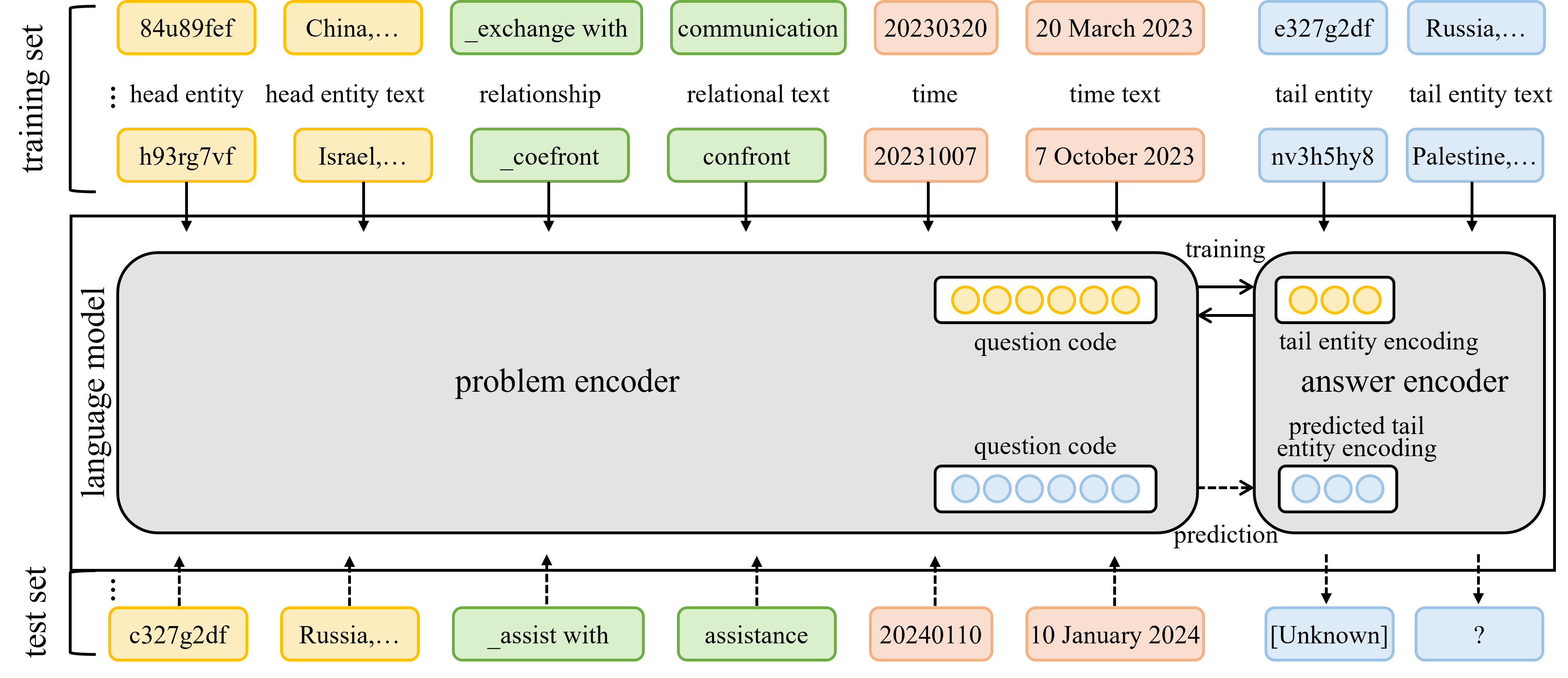}
  \caption{The training and prediction on quadruple knowledge with the large language model.}
  \label{fig:framework}
\end{figure*}

\section{Experimental Setup}

This section outlines the experimental setup, including the datasets, baseline models, and training configurations.

\subsection{Datasets}

We use two benchmark datasets for temporal knowledge graph reasoning: YAGO11k and WIKIdata12k. These datasets are commonly used for evaluating temporal reasoning models \citep{leblay2018deriving, garcia2020temporal}. YAGO11k is derived from Wikipedia, WordNet, and GeoNames \citep{mahdisoltani2015yago3}, while WIKIdata12k is constructed from Wikidata and Wikimedia Commons \citep{vrandevcic2014wikidata}. Both datasets support interpolation and extrapolation tasks.

\subsection{Baseline Models}

We compare our method against three baseline distillation models:

\begin{itemize}
    \item \textbf{BKD}: A traditional distillation technique that optimizes the Kullback--Leibler divergence to align the probability distributions produced by the teacher and the student models \citep{hinton2015distilling}.
    
    \item \textbf{FitNet}: An enhanced variant of basic distillation that applies supervision to intermediate hidden layers, enabling the student to capture richer internal representations from the teacher network \citep{romero2015fitnets}.
    
    \item \textbf{RKD}: A relation-aware distillation strategy that preserves structural information by aligning pairwise distances and angular relationships between teacher and student embedding spaces.
\end{itemize}

\begin{table}[t]
\centering
\caption{Statistical data of two datasets}
\label{tab:dataset_statistics}
\resizebox{\linewidth}{!}{
\begin{tabular}{lccccc}
\hline
Dataset & Entities & Relations & Train & Valid & Test \\
\hline
YAGO & 10{,}623 & 10 & 161{,}540 & 19{,}523 & 20{,}026 \\
WIKI & 12{,}544 & 24 & 539{,}286 & 67{,}538 & 63{,}110 \\
\hline
\end{tabular}
}
\end{table}

\begin{table}[t]
\centering
\caption{Introduction of classic TKG models}
\label{tab:tkg_models}
\resizebox{\linewidth}{!}{
\begin{tabular}{lcc}
\hline
Model & Encoder & Decoder \\
\hline
TTransE & -- & $\lVert s + p - o + t \rVert_2$ \\
TADistMult & $p_{\text{seq}} = \mathrm{LSTM}(p:t)$ & $(s \circ o)\, p_{\text{seq}}^{\top}$ \\
\hline
\end{tabular}
}
\end{table}

\subsection{Model Selection}

We adopt TTransE and TADistMult as the underlying knowledge graph embedding models. Both are widely used temporal KG methods and achieve competitive results on time-aware link prediction benchmarks \citep{leblay2018deriving, garcia2020temporal}. The encoder and decoder structures follow the settings described in their original papers.

In these formulations, $s, o \in \mathbb{R}^d$ denote subject and object embeddings, $p \in \mathbb{R}^d$ represents the relation embedding, and $t$ corresponds to the temporal representation. The operator $\circ$ indicates element-wise multiplication.

\begin{table*}[t]
\centering
\caption{Distillation results of different model method combinations on two datasets}
\label{tab:distillation_results}
\setlength{\tabcolsep}{4pt}
\resizebox{\textwidth}{!}{
\begin{tabular}{llccccc|ccccc}
\hline
 &  & \multicolumn{5}{c|}{YAGO} & \multicolumn{5}{c}{WIKI} \\
\cline{3-12}
Model & Method
& MRR & MR & Hits@1 & Hits@3 & Hits@10
& MRR & MR & Hits@1 & Hits@3 & Hits@10 \\
\hline
\multirow{4}{*}{TTransE}
& BKD
& \underline{7.65} & 1410.12 & 3.50 & \underline{7.83} & 15.61
& \textbf{7.94} & 2383.67 & \underline{4.75} & \underline{8.22} & 14.04 \\
& FitNet
& 7.59 & 1201.69 & 3.06 & 7.18 & \underline{16.48}
& 7.86 & 2148.86 & 3.93 & 7.78 & \underline{14.67} \\
& RKD
& 7.01 & \textbf{1186.27} & \underline{3.56} & 6.95 & 13.47
& 7.89 & \underline{2052.37} & 4.72 & 7.49 & 12.85 \\
& Ours
& \textbf{7.69} & \underline{1193.15} & \textbf{3.61} & \textbf{7.89} & \textbf{16.57}
& \underline{7.92} & \textbf{1985.63} & \textbf{4.86} & \textbf{8.36} & \textbf{14.94} \\
\hline
\multirow{4}{*}{TADistMult}
& BKD
& \textbf{61.90} & \underline{973.89} & \underline{58.51} & \underline{64.13} & \underline{67.59}
& 45.89 & 3150.11 & \underline{42.46} & 48.87 & 51.18 \\
& FitNet
& 58.44 & 986.92 & 54.71 & 60.29 & 65.34
& 43.92 & 3158.20 & 39.77 & 47.38 & 50.18 \\
& RKD
& 58.15 & 1089.57 & 54.48 & 61.72 & 65.17
& 42.72 & 3287.49 & 36.32 & 43.92 & 47.28 \\
& Ours
& \underline{61.87} & \textbf{965.35} & \textbf{58.73} & \textbf{64.15} & \textbf{67.68}
& \textbf{46.03} & 3142.85 & 42.50 & \textbf{49.16} & \underline{51.14} \\
\hline
\end{tabular}
}
\end{table*}

\subsection{Evaluation Metrics}

We assess model performance on two widely used temporal knowledge graph benchmarks. Evaluation is conducted using mean rank (MR), mean reciprocal rank (MRR), and Hits@1, Hits@3, and Hits@10, which are standard metrics for link prediction in knowledge graphs \citep{bordes2013translating, trouillon2016complex}. Hits@k reports the proportion of test queries for which the ground-truth entity is ranked within the top-$k$ predictions. A smaller MR indicates better overall ranking quality, while higher MRR and Hits@k values reflect stronger predictive accuracy and more reliable entity prioritization \citep{sun2019rotate}. These metrics collectively evaluate both global ranking performance and top-position precision.

\subsection{Training Settings}

All experiments are conducted on a workstation equipped with an Intel Core i9-10900K CPU and an NVIDIA RTX 3090 Ti GPU. Our implementation builds upon an extended version of the OpenKE framework \citep{han2018openke}, implemented in PyTorch.

To emphasize the knowledge transfer challenge, we intentionally create a substantial capacity gap between teacher and student models. The teacher embedding dimension is set to 400, whereas the student embedding size is restricted to 25. This setting reflects common practice in distillation studies for compact knowledge graph embeddings \citep{hinton2015distilling} and ensures that performance gains are not due to increased model capacity.

Training is performed with a batch size of 1024 for up to 10{,}000 epochs to ensure stable convergence. The temperature parameter used in soft distillation is fixed at 7, following standard practice in knowledge distillation literature. We employ the Adagrad optimizer to better handle sparse relational data and to adaptively adjust learning rates during optimization \citep{duchi2011adaptive}. All hyperparameters are selected based on validation performance.

\section{Results and Analysis}

\subsection{Experimental Results}

We evaluate the proposed framework on both the YAGO and WIKI datasets. Except for MR, all reported metrics are presented as percentages. Under identical experimental conditions, the best results are highlighted and the second-best values are underlined.

Across both datasets and backbone models, the student networks trained with our method consistently surpass those trained using conventional distillation strategies. Compared with BKD, improvements are observed on nearly all evaluation criteria. For TTransE on YAGO, performance increases are observed across MR, MRR, and Hits metrics, indicating better overall ranking and stronger top-position accuracy. Similar improvements are observed on the WIKI dataset, demonstrating robustness across different temporal knowledge distributions.

The TADistMult backbone shows comparable trends. When compared with FitNet and RKD, the proposed method achieves more stable and generally superior results across most metrics. In particular, noticeable gains on Hits@1 and Hits@3 suggest that the distilled student model ranks correct entities more accurately at higher positions, which is especially important in real-world temporal reasoning systems.

In a limited number of settings, certain baselines slightly outperform our approach on individual metrics. For instance, RKD exhibits marginally better MR results under specific configurations. These fluctuations may be attributed to differences in intermediate representation alignment or to the restricted capacity of the compact student model. Additional training epochs help reduce such gaps, suggesting that optimization stability plays a role.

\subsection{Ablation Analysis}

To further understand the impact of LLM-guided supervision, we perform ablation studies by removing the LLM-related distillation component. Compared with the BKD baseline, incorporating semantic signals from large language models consistently improves performance across most metrics on both datasets. The results indicate that structured knowledge extracted from LLMs provides complementary information beyond task-specific supervision, enhancing the generalization ability of lightweight students.

\section{Conclusion}

We present an LLM-augmented distillation framework for temporal knowledge graph reasoning designed for efficient deployment in resource-constrained environments. By transferring structured temporal knowledge from high-capacity teacher models and leveraging semantic signals from large language models, the proposed approach improves predictive accuracy while preserving a compact model size. Extensive experiments on multiple datasets and backbone architectures demonstrate consistent gains over traditional distillation methods, validating both the effectiveness and practicality of the proposed framework.

\bibliography{custom}

\end{document}